\documentclass[runningheads]{llncs}

\usepackage[T1]{fontenc}

\usepackage{graphicx}
\usepackage{hyperref}
\usepackage{url}
\usepackage{subfig}
\usepackage{multirow}
\usepackage{amsmath}
\usepackage{amssymb}
\usepackage{mathtools}

\newcommand{\textbfp}[1]{\vspace{0.5em}\noindent\textbf{#1}}
\newcommand\attr{\mathbf{\hat{a}}}
\newcommand\steerf{\mathbf{\hat{f}}}
\newcommand\steerm{\mathbf{\hat{m}}}

\usepackage{color}

\newcommand\extrafootertext[1]{%
    \bgroup
    \renewcommand\thefootnote{\fnsymbol{footnote}}%
    \renewcommand\thempfootnote{\fnsymbol{mpfootnote}}%
    \footnotetext[0]{#1}%
    \egroup
}

\begin{document}

\title{Learning Lagrangian Fluid Mechanics with E($3$)-Equivariant Graph Neural Networks}

\titlerunning{Learning Lagrangian Fluid Mechanics with E($3$)-Equivariant GNNs}

\author{Artur P. Toshev\inst{1} \and
Gianluca Galletti\inst{1} \and
Johannes Brandstetter\inst{2} \and
Stefan Adami\inst{1} \and
Nikolaus A. Adams\inst{1}
}
\authorrunning{A. P. Toshev et al.}

\institute{Technical University of Munich, School of Engineering and Design, Chair of Aerodynamics and Fluid Mechanics, Garching, Germany \\
\email{\{artur.toshev,g.galletti\}@tum.de}\\
\and
Microsoft Research AI4Science}

\maketitle

\begin{abstract}
We contribute to the vastly growing field of machine learning for engineering systems by demonstrating that equivariant graph neural networks have the potential to learn more accurate dynamic-interaction models than their non-equivariant counterparts. We benchmark two well-studied fluid-flow systems, namely 3D decaying Taylor-Green vortex and 3D reverse Poiseuille flow, and evaluate the models based on different performance measures, such as kinetic energy or Sinkhorn distance. In addition, we investigate different embedding methods of physical-information histories for equivariant models. We find that while currently being rather slow to train and evaluate, equivariant models with our proposed history embeddings learn more accurate physical interactions. 

\extrafootertext{Our code will be released under \url{https://github.com/tumaer/sph-hae}
}

\end{abstract}

\begin{figure}[ht]
  \centering
  \begin{minipage}[c]{0.45\textwidth}
    \centering
    \subfloat[\label{subfig:a} Reverse Poiseuille flow (RPF)]{
        \includegraphics[width=0.3\textwidth]{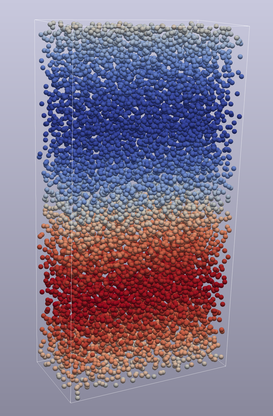}%
        \includegraphics[width=0.3\textwidth]{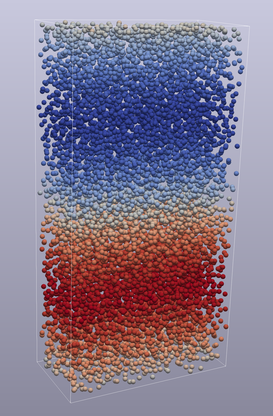}%
        \includegraphics[width=0.3\textwidth]{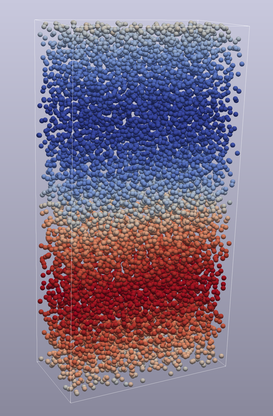}} \\
        \vspace{-6px}
    \subfloat[\label{subfig:b} Taylor-Green vortex (TGV)]{
        \includegraphics[width=0.3\textwidth]{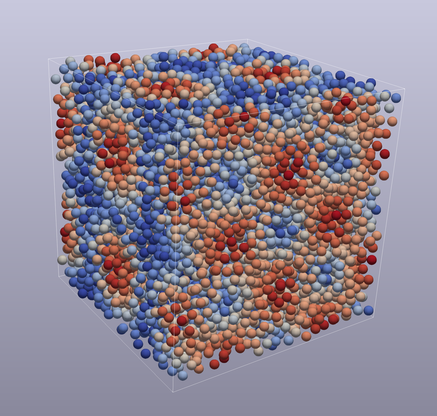}%
        \includegraphics[width=0.3\textwidth]{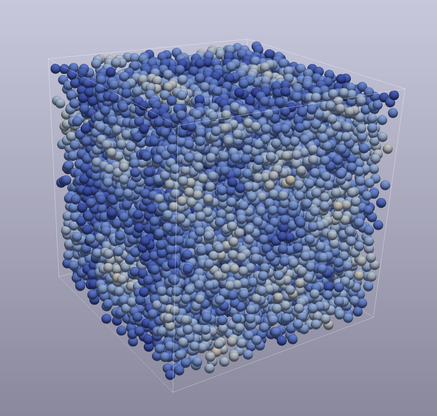}%
        \includegraphics[width=0.3\textwidth]{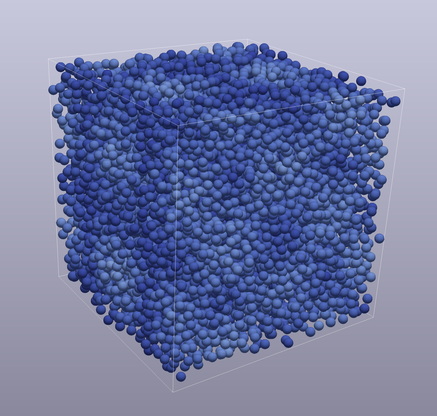}}
  \end{minipage}
  \begin{minipage}[c]{0.45\textwidth}
    \centering
    \subfloat[\label{subfig:c} Total kinetic-energy]{\includegraphics[width=0.98\textwidth]{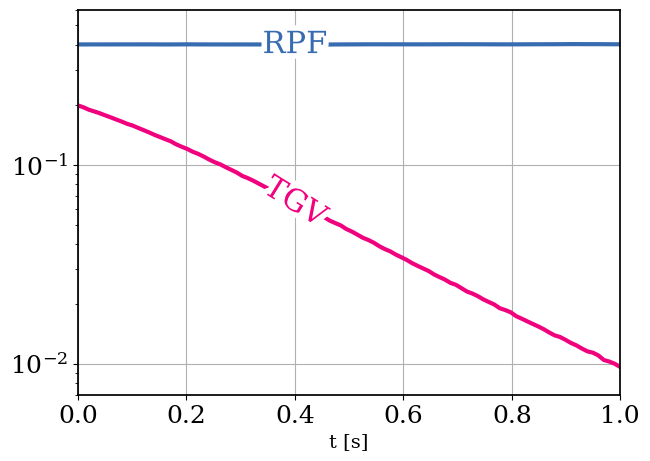}} 
  \end{minipage}
  \captionsetup{justification=centering}
  \caption{Time snapshots of x-velocity of reverse Poiseuille flow (a), velocity magnitude of Taylor-Green vortex flow (b), and kinetic-energy evolution (c).}%
  \label{fig:main}%
\end{figure}

\section{Particle-based fluid mechanics}

The Navier-Stokes equations (NSE) are omnipresent in fluid mechanics.
However, for the majority of problems, solutions are analytically intractable, and obtaining accurate solutions necessitates numerical approximations.
Those can be split into two categories: grid/mesh-based (Eulerian description) and particle-based (Lagrangian description). 

\textbfp{Smoothed Particle Hydrodynamics. } 
In this work, we investigate Lagrangian methods, more precisely the Smoothed Particle Hydrodynamics (SPH) approach, which was independently developed by \cite{gingold1977smoothed} and \cite{lucy1977numerical} to simulate astrophysical systems. Since then, SPH has established itself as the preferred approach in various applications ranging from free-surface flows such as ocean waves \cite{violeau2016smoothed} 
to selective laser melting in additive manufacturing \cite{weirather2019smoothed}. The main idea behind SPH is to represent fluid properties at discrete points in space and to use truncated radial interpolation kernel functions to approximate them at any arbitrary location.
The kernel functions can be interpreted as state-statistics estimators which define continuum-scale interactions between particles. 
The justification for truncating the kernel support is the assumption of the locality of interactions between particles. The resulting discretized equations are integrated in time using numerical integration techniques such as the symplectic Euler scheme, by which the particle positions are updated.
 
To generate training data for our machine learning tasks, we implemented our own fully-differentiable SPH solver in JAX \cite{jax2018github} based on the transport velocity formulation of SPH by \cite{adami2013transport}, which achieves homogeneous particle distributions over the domain. We then selected two flow cases,
which have been extensively studied in fluid mechanics: 3D Taylor-Green vortex and 3D reverse Poiseuille flow. 
We expect to open-source the datasets in the near future.

\textbfp{Taylor-Green Vortex. }
The Taylor-Green vortex system (TGV, see Figure~\ref{fig:main} (a)) was introduced by Taylor \& Green in 1937 to study turbulence \cite{taylor1937mechanism}.
We investigate the TGV with Reynolds number of $\text{Re}=100$, which is neither laminar nor turbulent, i.e. there is no layering of the flow (typical for laminar flows), but also the small scales caused by vortex stretching do not lead to a fully developed energy cascade (typical for turbulent flows) \cite{brachet1984taylor}. 
We compute the Reynolds number $Re=UL/\eta$ as in \cite{adami2013transport} with domain size $L=1$, reference velocity $U=1$, and dynamic viscosity $\eta=0.01$. We note that this setup differs from the one in \cite{brachet1984taylor}, where the domain is $L=2\pi$. 
We use the initial velocity field from \cite{brachet1984taylor}:
\begin{subequations}
\begin{align}
    u &= \sin(k x) \cos(k y) \cos(k z) \ , \\
    v &= - \cos(k x) \sin(k y) \cos(k z) \ ,  \\
    w &= 0 \ ,
\end{align}
\end{subequations}
 where $k=2 \pi / L$.
The TGV dataset used in this work consists of 80/10/10 trajectories for training/validation/testing,
where each trajectory comprises 8000 particles. Each trajectory spans $1$s physical time and was simulated with $dt=0.001$s starting from a random initial particle distribution. We choose to train the learned solver on 10x larger time steps, i.e. temporal coarsening, which we implement by subsampling every 10th frame resulting in 100 samples per trajectory.

\textbfp{Reverse Poiseuille Flow. }
The Poiseuille flow, i.e. laminar channel flow, is another well-studied fluid mechanics problem. However, the channel flow requires the treatment of wall-boundary conditions, which is beyond the focus of the current work. Therefore, in this work, we consider data obtained by reverse Poiseuille flow (RPF, see Figure~\ref{fig:main} (b)) \cite{fedosov2008reverse}, which essentially consists of two opposing streams in a fully periodic domain.
In terms of the SPH implementation, the flow is exposed to opposite force fields, i.e. the upper and lower half are accelerated in negative $x$ direction and positive $x$ direction, respectively.
Here we also choose to work with $\text{Re}=100$, in which case the flow is not purely laminar and there is no analytical solution for the velocity profile. 
The domain has size 1/2/0.5 in x/y/z directions (width, height, depth), and for the computation of the Reynolds number $Re=UL/\eta$ we use $U=1$, $L=1$, $\eta=0.01$.

Due to the statistically stationary \cite{pope2000turbulent} solution of the flow, the RPF dataset consists of one long trajectory spanning 100s. The flow field is again discretized by 8000 particles and simulated with $dt=0.001$, followed by subsampling at every 10th step.
Thus, we again aim to train models to perform temporal coarsening. The resulting number of training/validation/testing instances is the same as for TGV, namely 8000/1000/1000.

\section{(Equivariant) graph network-based simulators}
\label{sec:2}
We first formalize the task of autoregressive prediction of the next state of a Lagrangian flow field based on the notation from \cite{sanchez2020learning}. If $X^t$ denotes the state of a particle system at time $t$, one full trajectory of $K+1$ steps can be written as $\textbf{X}^{t_{0:K}}=(\textbf{X}^{t_0}, \ldots, \textbf{X}^{t_K})$. Each state $\textbf{X}^t$ is made up of $N$ particles, namely $\textbf{X}^t = (\textbf{x}_1^t, \textbf{x}_2^t, \dots \textbf{x}_N^t)$, where each $\textbf{x}_i$ is the state vector of the $i$-th particle. 
However, the inputs to the learned simulator can span multiple time instances. Each node $\textbf{x}_i^t$ can contain node-level information like the current position $\mathbf{p}^t_i$ and a time sequence of $H$ previous velocity vectors $\dot{\textbf{p}}^{t_{k-H:k}}$, as well as global features like the external force vector $\mathbf{F}_i$ in the reverse Poiseuille flow. To build the connectivity graph, we use an interaction radius of $\sim 1.5$ times the average interparticle distance, which results in around 10-20 one-hop neighbors.

\textbfp{Graph Network-based Simulator.}
The GNS framework~\cite{sanchez2020learning} is one of the most popular learned surrogates for engineering particle-based simulations. The main idea of the GNS model is to use the established encoder-processor-decoder architecture~\cite{battaglia2018relational} with a processor that stacks several message passing layers \cite{gilmer2017neural}. One major strength of the GNS model lies in its simplicity given that all its building blocks are regular MLPs. 
However, the performance of GNS when predicting long trajectories strongly depends on the choice of Gaussian noise to perturb the input data.
Additionally, GNS and other non-equivariant models are less data-efficient~\cite{batzner2022e3}. For these reasons, we implement and tune GNS as a comparison baseline, and employ it as an inspiration for which setup, features, and hyperparameters to use for equivariant models.

\textbfp{Steerable E(3)-equivariant Graph Neural Network. }
SEGNNs~\cite{brandstetter2021segnn} are an instance of E($3$)-equivariant GNNs, i.e. GNNs that are equivariant with respect to isometries of the Euclidean space (rotations, translations, and reflections). Most E($3$)-equivariant GNNs tailored for prediction of molecular properties~\cite{thomas2018tensor,batzner2022e3,batatia2022mace} parametrize Clebsch-Gordan tensor products using a learned embedding of pairwise distances.
In contrast, the SEGNN model uses general steerable node and edge attributes ($\attr_i$ and $\attr_{ij}$ respectively) to condition the layers directly. In particular, SEGNNs introduce the concept of steerable MLPs, which are linear Clebsch-Gordan tensor products $\otimes_{CG}^{\mathcal{W}_\attr}$ parametrized by learnable parameters $\mathcal{W}_\attr$ interleaved with gated non-linearities $\sigma$ \cite{weiler2018gated}. Namely, the hidden state $\steerf$ at layer $l+1$ is updated as
\begin{align}
    \steerf^{l+1} := \sigma(\mathcal{W}_{\attr} \steerf^l) \quad \text{with} \quad \mathcal{W}_{\attr} \steerf := \steerf \otimes_{CG}^{\mathcal{W}_\attr} \attr \ .
    \label{eq:layer}
\end{align}
Due to these design choices, SEGNNs are well suited for a wide range of engineering problems, where various vector-valued features need to be modeled in an E($3$) equivariant way. In practice, SEGNNs extend upon the message passing paradigm \cite{gilmer2017neural} using steerable MLPs of Equation (\ref{eq:layer}) for both message $\textit{M}_{\attr_{ij}}$ and node update functions $\textit{U}_{\attr_i}$. The $i$-th node steerable features $\steerf_i$ are updated as
\begin{align}
    \steerm_{ij} &= \textit{M}_{\attr_{ij}}\left( \steerf_i, \steerf_j, \| x_i - x_j \|^2 \right), \label{eq:mp_message} \\
    \steerf^{\prime}_i &= \textit{U}_{\attr_i}\left( \steerf_i, \sum_{j\in\mathcal{N}(i)} \steerm_{ij} \right), \label{eq:mp_update}
\end{align}
where $\mathcal{N}(i)$ is the neighborhood of node $i$. In Equation (\ref{eq:mp_message}), $\textit{M}_{\attr_{ij}}$ has the subscript $\attr_{ij}$ because it is conditioned on the edge attributes, whereas $\textit{U}_{\attr_i}$ is conditioned on the node attributes $\attr_i$.

\begin{figure}[t]
    \centering
    \subfloat[\label{subfig:att} Layer-wise attribute embeddings]{
        \includegraphics[height=3.6cm]{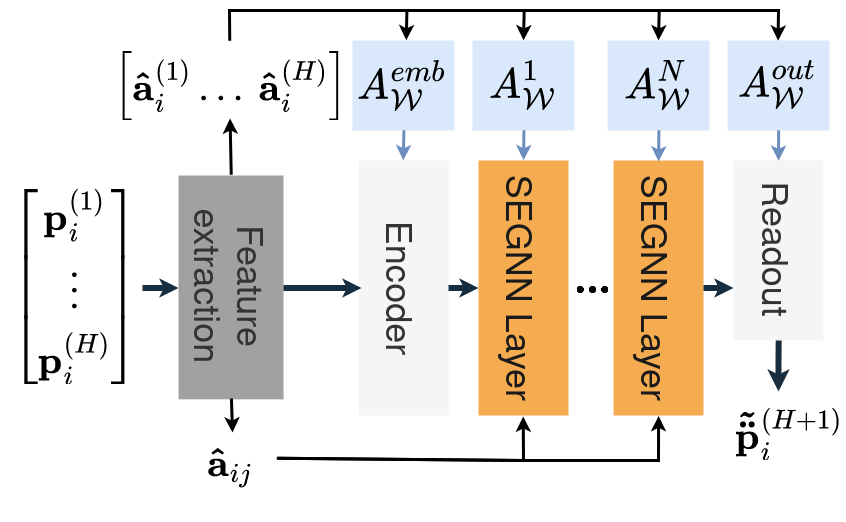}
    }%
    \hspace{2pt}
    \subfloat[\label{subfig:vels} Examples of embedding effects]{
        \includegraphics[height=3.6cm]{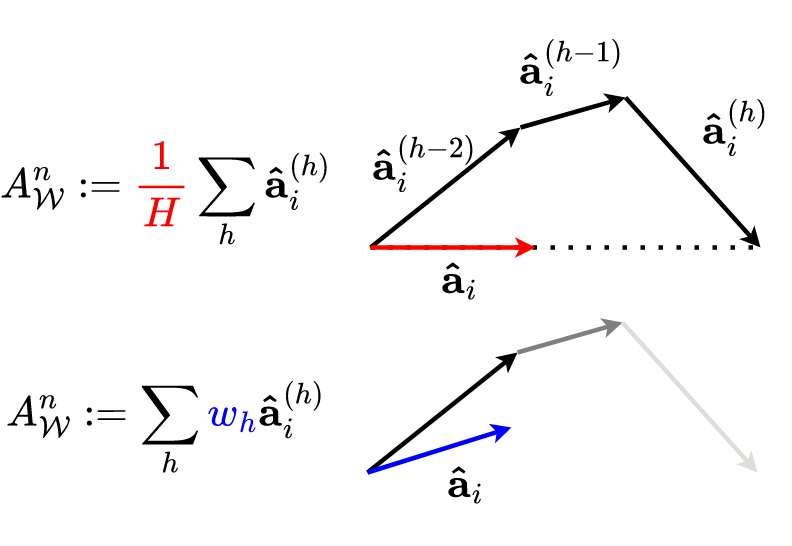}
    }
    \caption{SEGNN architecture with Historical Attribute Embedding.
    (a): $A_{\mathcal{W}}^n$ is a learnable embedding of previous velocities for node attributes $\attr_i^{(h)}$. (b): effects of $A_{\mathcal{W}}^n$ as arithmetic mean (top) and weighted mean of past attributes (bottom).}%
    \label{fig:architecture}%
\end{figure}

\textbfp{Historical Attribute Embedding (HAE).} Finding physically meaningful edge and node attributes is crucial for good performance since every Clebsch-Gordan tensor product is conditioned on them. For the problems at hand, we empirically found that a strong choice for steerable attributes is
\begin{align}
    \attr_{ij} &=  Y\left(\textbf{p}_{i}-\textbf{p}_{j}\right), \\ 
    \attr_i &= A_{\mathcal{W}} \left( Y\left(\dot{\textbf{p}}_i^{(1:H)}\right) + \sum_{j\in \mathcal{N}(i)} \attr_{ij} \right) = A_{\mathcal{W}} \left( \attr_i^{(1:H)} \right),
\end{align}
where $\attr_i$ and $\attr_{ij}$ are the node and edge attributes respectively, $Y_m^{(l)}:S^2\rightarrow \mathbb{R}$ is the spherical harmonics embedding, and $A_{\mathcal{W}}$ is a function parameterized by $\mathcal{W}$ that embeds the historical node attributes $\attr_i^{(h)}$.
In particular, we investigate the averaging $A_{\mathcal{W}, avg}$, weighted averaging $A_{\mathcal{W}, lin}$, and steerable MLP embedding conditioned on the most recent velocity  $A_{\mathcal{W}, \otimes}$. 
\begin{align}
    A_{\mathcal{W}, avg} &:= \frac{1}{H}\sum_{h} \attr_i^{(h)} \ , \\  
    A_{\mathcal{W}, lin} &:= \sum_{h} w_h\attr_i^{(h)} \ , \\
    A_{\mathcal{W}, \otimes} &:= \sigma \left( \attr_i^{(1:H)} \otimes_{CG}^{\mathcal{W}} \attr_i^{H} \right) \ .
\end{align}
Figure \ref{fig:architecture} sketches the HAE-SEGNN architecture.
Subfigure \ref{subfig:att} connects past particle positions $\mathbf{p}^{(1:H)}$ and their embeddings $A_{\mathcal{W}}^n$ within the updated architecture, whereas subfigure \ref{subfig:vels} shows the effect of $A_{\mathcal{W}, avg}$ and $A_{\mathcal{W}, lin}$.

We found that initializing the embedding weights for $A_{\mathcal{W}, lin}$ and $A_{\mathcal{W}, \otimes}$ with $N(\mu=\frac{1}{\#\mathcal{W}}, \sigma=\frac{1}{\sqrt{fan_{in}}})$ 
(shifted initialization to resemble the average $A_{\mathcal{W}, avg}$)
makes training quicker and also slightly improves final performance. This behavior reiterates the significance of attributes in conditioning the architecture, and the inclusion of potentially less relevant attributes can lead to a substantial decrease in performance. 

We currently don't have a systematic way of finding the attributes, but we found that in engineering systems having physical features (such as velocity and force) in the attributes has a positive effect, as one could see it as conditioning the network on system dynamics.
Exploring an algorithmic framework for finding attributes in the broader context was not investigated and is left to future work.

\textbfp{Related work. }
Related steerable E($3$)-equivariant GNNs, such as ~\cite{thomas2018tensor,batzner2022e3,batatia2022mace,pezzicoli2022se} are mostly tailored towards molecular property prediction tasks, and thus restrict the parametrization of tensor products to an MLP-embedding of pairwise distances.
This is a reasonable design choice since distances are crucial information for molecules, but not straightforward to adapt to fluid dynamics problems where prevalent quantities are e.g. force and momentum.
Another family of E($3$)-equivariant GNNs are models that use invariant quantities, such as distances and angles ~\cite{satorras2021en,gasteiger2020directional,gasteiger2021gemnet}. Although these models have an advantage concerning runtimes since no Clebsch-Gordan tensor product is needed, they cannot a priori model vector-valued information in an E($3$) equivariant way.  
On a slightly more distant note, there has been a rapid raise in physics-informed neural networks (PINNs)~\cite{raissi2019physics} and equivariant counterparts thereof~\cite{lagrave2022equivariant}, as well as operator learning~\cite{lu2019deeponet,li2020neural,li2021fourier,gupta2022towards}
, where functions or surrogates are learned in an Eulerian (grid-based) way. Furthermore, equivariant models have been applied to grid-based data \cite{wang2021incorporating,lagrave2022equivariant} utilizing group-equivariant CNNs \cite{cohen2016gcnn,weiler20183d}.
Recently, Clifford algebra-based layers \cite{brandstetter2022clifford} have been proposed on grids as well as graph-structured data \cite{ruhe2023geometric,ruhe2023clifford}, but exploring their performance on SPH data is left to future work. Non-equivariant deep learning surrogates for Lagrangian dynamics were introduced for particles~\cite{sanchez2020learning}, meshes~\cite{pfaff2020learning}, and within complex geometries~\cite{mayr2021boundary}.
    
\section{Results}
The task we train on is the autoregressive prediction of accelerations $\ddot{\mathbf{p}}$ given the current position $\mathbf{p}_i$ and $H=5$ past velocities of the particles $\dot{\textbf{p}}_i^{(1:H)}$. 
The influence of the choice of $H$ is discussed in detail in the supplementary materials to \cite{sanchez2020learning} and we use the same value as suggested in this paper.
For training SEGNNs, we verified that adding Gaussian noise to the inputs~\cite{sanchez2020learning} indeed significantly improves performance. In addition, we train both models by employing the pushforward trick \cite{brandstetter2022message} with up to five pushforward steps and an exponentially decaying probability with regard to the number of steps.
We measured the performance of the GNS and the SEGNN models in four aspects when evaluating on the test datasets:
\begin{enumerate}
    \item \textit{Mean-squared error} (MSE) of particle positions $\text{MSE}_p$ when rolling out a trajectory over 100 time steps (1 physical second for both flow cases). This is also the validation loss during training. 
    \item \textit{Sinkhorn distance} as an optimal transport distance measure between particle distributions. Lower values indicate that the particle distribution is closer to the reference one. 
    \item \textit{Kinetic energy} $E_{kin}$ ($=0.5 m v^2$) as a global measure of physical behavior.
\end{enumerate}

\begin{table}[ht]
\centering
\setlength{\tabcolsep}{8pt}
\renewcommand{\arraystretch}{1.3}
\caption{Performance measures on the Taylor-Green vortex and reverse Poiseuille flow.
The Sinkhorn distance is averaged over test rollouts.}
\begin{tabular}{cccc|ccc}
 & \multicolumn{3}{c|}{Taylor-Green vortex} & \multicolumn{3}{c}{reverse Poiseuille flow} \\ \cline{2-7}
\multicolumn{1}{c}{} & \multicolumn{1}{c}{MSE$_p$} & \multicolumn{1}{c}{$\text{MSE}_{E_{kin}}$} & $\overline{\text{Sinkhorn}}$ & \multicolumn{1}{c}{MSE$_p$} & \multicolumn{1}{c}{$\text{MSE}_{E_{kin}}$} & $\overline{\text{Sinkhorn}}$ \\ 
\hline
\multicolumn{1}{c|}{GNS} & 6.7e-6 & 7.1e-3 & 1.2e-7 & 1.4e-6 & 2.2e-2 & 4.1e-7 \\
\multicolumn{1}{c|}{SEGNN$_{avg}$} & 1.6e-6 &  8.4e-3 & 2.9e-8 &  1.4e-6 & 8.2e-3 & 1.4e-7 \\
\multicolumn{1}{c|}{SEGNN$_{lin}$} & \textbf{1.4e-6} & \textbf{3.1e-4} & 2.0e-8 & \textbf{1.3e-6} & 2.0e-2 & 1.2e-7 \\
\multicolumn{1}{c|}{SEGNN$_{\otimes}$} & \textbf{1.4e-6} & 1.9e-3 & \textbf{1.6e-8} & \textbf{1.3e-6} & \textbf{9.4e-4} & \textbf{9.1e-8}
\end{tabular}
\label{table:comparison}
\end{table}

Performance comparisons are summarized in Table \ref{table:comparison}. GNS and SEGNN have 1.2M and 360k parameters respectively for both Taylor-Green and reverse Poiseuille (both have 10 layers, but 128 vs 64-dim features). 
For all SEGNN models, we used maximum spherical harmonics order $l_{max}=1$ attributes as well as features; we found that in our particular case, higher orders become computationally unfeasible to train and evaluate. With regards to runtime for 8000 particles dynamics, the GNS model takes around 35ms per step, all SEGNN models take roughly 150ms per step, and the original SPH solver takes around 100ms per 10 steps (note: we learn to predict every 10th step). Both our SPH solver as well as GNS and SEGNN models are implemented in the Python library JAX \cite{jax2018github}, and both use the same neighbors-search implementation from the JAX-MD library \cite{jaxmd2020}, making for a fair runtime comparison. 
It is known that steerable equivariant models are slower than non-equivariant ones, which is related to how the Clebsch-Gordan tensor product is implemented on accelerators like GPUs. However, we observed that equivariant models reach their peak performance with fewer parameters, and they often significantly outperform GNS, especially when measuring physics quantities like kinetic energy or Sinkhorn distances.

\begin{figure}[ht]%
    \centering
    \includegraphics[clip, trim=-1.3cm 5.0cm -2cm 0cm,width=0.94\textwidth]{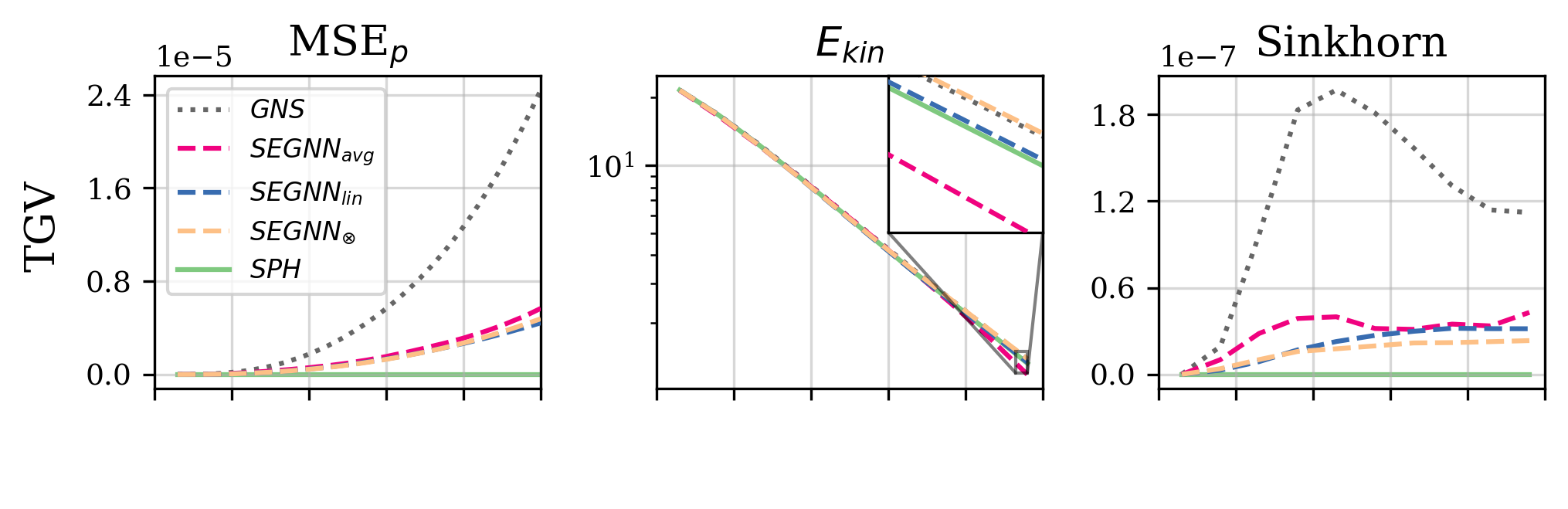}\\ 
    \hspace{8pt} \includegraphics[clip, trim=0.0cm 1.cm -2cm 0cm,width=0.94\textwidth]{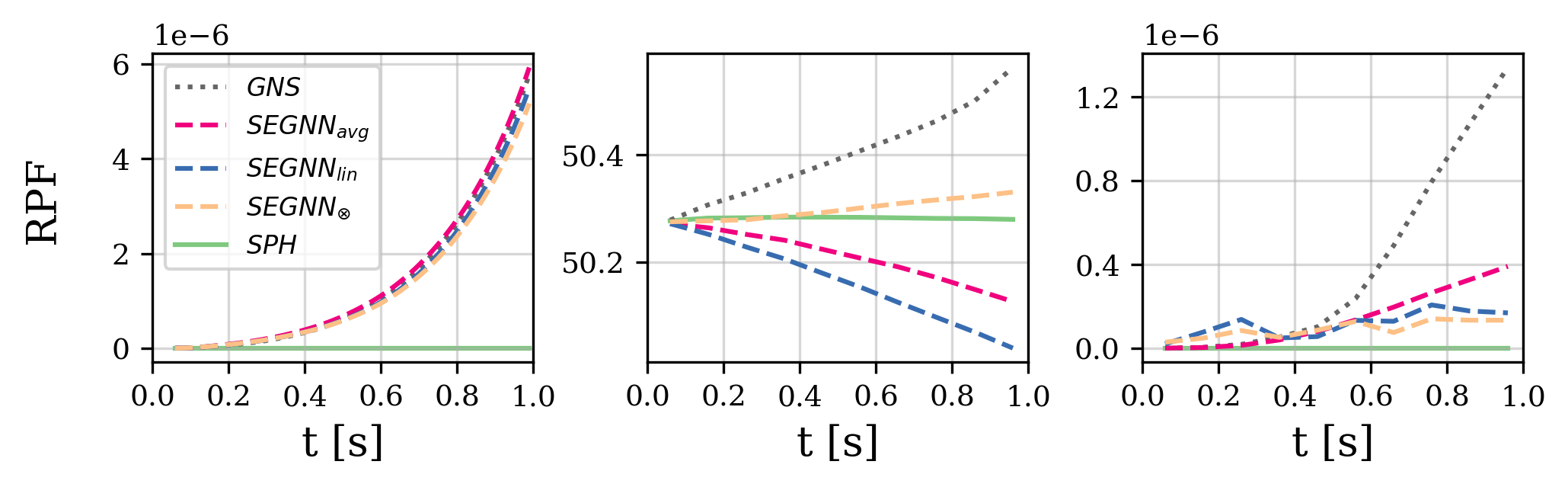}\\ 
    \caption{Evolution of performance measures over time on the Taylor-Green vortex (top) and reverse Poiseuille flow (bottom).}%
    \label{fig:metrics}%
\end{figure}

\textbfp{Taylor-Green Vortex. } 
One of the major challenges of the Taylor-Green dataset is the varying input and output scales throughout a trajectory, in our case by up to one order of magnitude. This results in the larger importance of initial frames in the loss even after data normalization. 
Figure \ref{fig:metrics} (top) summarizes the performance properties of the Taylor-Green vortex experiment.
Both models are able to match the ground truth kinetic energy. However, all SEGNN models achieve 20 times lower MSE$_{E_{kin}}$ errors, and regarding MSE$_p$, GNS predictions drift away from the reference SPH trajectory much earlier. 

\textbfp{Reverse Poiseuille Flow. }
The challenge of the reverse Poiseuille case lies in the different velocity scales between the main flow direction ($x$-axis) and the $y$ and $z$ components of the velocity. In contrast to GNS, whose inputs we can normalize with the direction-dependent dataset statistics, this breaks equivariance and we are forced to normalize the SEGNN inputs only in magnitude. Although such unbalanced velocities are used as inputs, target accelerations in $x$-, $y$-, and $z$-direction all have similar distributions. This, combined with temporal coarsening makes the problem sensitive to input deviations.
Additionally, including the external force vector $\mathbf{F}_i$ to either the node features or SEGNN attributes has a positive impact on the results.
Figure \ref{fig:metrics} (bottom) shows that SEGNNs reproduce the particle distributions quite well, whereas GNS show signs of particle-clustering artifacts, leading to a much larger Sinkhorn distance.

\section{Future Work}
In this work, we demonstrate that equivariant models are well suited to capture the underlying physical properties of particle-based fluid mechanics systems. We found that conditioning on physical quantities through our tensor product historical embedding increases expressive power at almost no additional cost. Moreover, employing more recent training strategies, such as the pushforward trick, has proven to be helpful in stabilizing training and improving performance.
Finally, selecting suitable (physical) performance measures different than plain MSE errors is crucial for assessing and improving deep learning models.

Interesting directions for future work include accelerating the inference time of equivariant GNNs as well as developing more specialized and expressive equivariant building blocks. 
We conjecture that together with such extensions, equivariant models offer a promising direction to tackle some of the long-standing problems in fluid mechanics, such as the learning of coarse-grained representations of turbulent flow problems, e.g. Taylor-Green \cite{brachet1984taylor}, or learning the multi-resolution dynamics of NSE problems \cite{hu2017consistent}. 

\section*{Acknowledgements}

We are thankful to the developers of the \texttt{e$3$nn-jax} library \cite{e3nn_paper}, which offers efficient implementations of E($3$)-equivariant building blocks.

\bibliographystyle{splncs04}
\bibliography{gsi2023}

\end{document}